\documentclass{article}

\usepackage{arxiv}

\usepackage[utf8]{inputenc} % allow utf-8 input
\usepackage[T1]{fontenc}    % use 8-bit T1 fonts
\usepackage{hyperref}       % hyperlinks
\usepackage{url}            % simple URL typesetting
\usepackage{booktabs}       % professional-quality tables
\usepackage{nicefrac}       % compact symbols for 1/2, etc.
\usepackage{microtype}      % microtypography
\usepackage{lipsum}		% Can be removed after putting your text content
\usepackage{graphicx}
\usepackage{natbib}
\usepackage{doi}

\usepackage{mathtools}
\usepackage{accents}
\usepackage{subfig}
\usepackage{multirow}

\title{NFISiS: New Perspectives on Fuzzy Inference Systems for Renewable Energy Forecasting}

\author{ \href{https://orcid.org/0000-0002-3258-0025}{\includegraphics[scale=0.06]{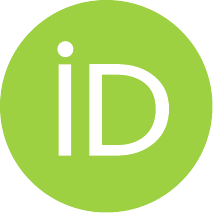}\hspace{1mm}Kaike Sa Teles Rocha Alves}\\%\thanks{Use footnote for providing further
	Computational Modeling \\
	Federal University of Juiz de Fora\\
	Juiz de Fora, MG \\
	\texttt{kaike.alves@outlook.com} \\
	\And
	\href{https://orcid.org/0000-0001-7458-8976}{\includegraphics[scale=0.06]{orcid.pdf}\hspace{1mm}Eduardo Pestana de Aguiar} \\
	Department of Industrial and Mechanical Engineering\\
	Federal University of Juiz de Fora\\
	Juiz de Fora, MG \\
	\texttt{eduardo.aguiar@ufjf.br} \\
}

\renewcommand{\shorttitle}{\textit{arXiv} Template}

\hypersetup{
pdftitle={A template for the arxiv style},
pdfsubject={Machine Learning},
pdfauthor={Kaike Sa Teles Rocha Alves},
pdfkeywords={Fuzzy inference systems, time series forecasting, feature selection, genetic algorithm, ensemble},
}

\begin{document}
\maketitle

\begin{abstract}
	Deep learning models, despite their popularity, face challenges such as long training times and a lack of interpretability. In contrast, fuzzy inference systems offer a balance of accuracy and transparency. This paper addresses the limitations of traditional Takagi-Sugeno-Kang fuzzy models by extending the recently proposed New Takagi-Sugeno-Kang model to a new Mamdani-based regressor. These models are data-driven, allowing users to define the number of rules to balance accuracy and interpretability. To handle the complexity of large datasets, this research integrates wrapper and ensemble techniques. A Genetic Algorithm is used as a wrapper for feature selection, creating genetic versions of the models. Furthermore, ensemble models, including the Random New Mamdani Regressor, Random New Takagi-Sugeno-Kang, and Random Forest New Takagi-Sugeno-Kang, are introduced to improve robustness. The proposed models are validated on photovoltaic energy forecasting datasets, a critical application due to the intermittent nature of solar power. Results demonstrate that the genetic and ensemble fuzzy models, particularly the Genetic New Takagi-Sugeno-Kang and Random Forest New Takagi-Sugeno-Kang, achieve superior performance. They often outperform both traditional machine learning and deep learning models while providing a simpler and more interpretable rule-based structure. The models are available online in a library called nfisis (\url{https://pypi.org/project/nfisis/}).
\end{abstract}

% keywords can be removed

%---------------------
\section{Introduction}
%---------------------

Deep learning (DL) techniques have become increasingly popular due to technological advances and the growing complexity of databases. However, DL models are more intricate and susceptible to several issues, such as vanishing gradients, overfitting, long training times\cite{shrestha2019review}, and the inability to provide interpretable results due to their black-box nature \cite{wang2020recent, rai2020explainable}. In contrast, fuzzy inference systems (FIS), represent a class of machine learning models that combine accuracy with interpretability. These systems are primarily divided into two categories: Mamdani \cite{mamdani1974application} and Takagi-Sugeno-Kang (TSK) \cite{takagi1985fuzzy, sugeno1988structure}. These systems have been successfully applied in various fields, including finance, business, and management \cite{bojadziev2007fuzzy}, medicine \cite{szczepaniak2012fuzzy}, engineering \cite{precup2011survey}, chemistry \cite{komiyama2017chemistry}, pattern recognition \cite{melin2011face}, and fault detection \cite{lemos2013adaptive}. However, Alves et al. \cite{alves2024new, sa2024financial} identified several limitations in TSK models, including: i) a lack of a unique approach for defining TSK rules; ii) data-driven TSK models are usually more complex, have many hyper-parameters, and involve hybridizations; iii) there is no direct control over the number of rules in data-driven approaches; and iv) the consequent part of TSK models lacks explainability, as it consists of polynomial functions. To address these limitations, the authors proposed the New Takagi-Sugeno-Kang (NTSK) model.

NTSK is a data-driven fuzzy model with reduced complexity and increased interpretability. Alves et al. also highlight the following: first, the number of rules is a hyper-parameter, allowing the user to set the exact number of rules that balance the trade-off between accuracy and interpretability. Second, the model defines the rules based on the tangent of the target value. This additional information enhances the interpretability of the consequent part. Finally, the approach has a reduced number of hyper-parameters and lower complexity, making it suitable for use in various real-world applications. The simulations demonstrated the superior performance of the proposed model, surpassing the performance of more complex models, such as DL.

In this paper, the concepts of NTSK will be extended to a new Mamdani fuzzy inference system (NMFIS) called the new Mamdani regressor (NMR). NMR is an autonomous model that defines the rule-based structure from data using a data-driven approach. The only requirement is to specify the desired number of rules. Like NTSK, NMR automatically defines the fuzzy sets and rules, offering a simple structure and accurate results, making it suitable for implementation in real-world applications. Nevertheless, with the growing complexity of databases, one challenge remains: which attributes should be selected to optimize the model's performance while considering the trade-off between accuracy and interpretability? The answer to this question depends on feature selection techniques, which can be divided into five main types: filter, wrapper, embedded, hybrid, and ensemble \cite{zebari2020comprehensive}. 

Filter methods are a statistic-based approach, but they neglect the integration between the selected subset and the performance of the learning algorithm \cite{eesa2015novel}. In contrast, wrapper methods use a metric that measures the learning algorithm's performance to identify the feature set that leads to the best results. However, since it is infeasible to test all possible combinations of features, heuristic and metaheuristic approaches are employed, such as randomized search \cite{mao2019wrapper}, genetic algorithm (GA) \cite{sohail2023genetic}, and the ant colony optimization \cite{forsati2014enriched}. In embedded methods, the feature selection technique is integrated into the learning algorithm by adjusting the model's internal parameters. Decision trees, random forests, and gradient boosting are examples of embedded feature selection techniques \cite{guyon2003introduction}. Hybrid methods combine multiple feature selection approaches in a multi-step process \cite{hsu2011hybrid}. Finally, an ensemble learner implements multiple weak learning algorithms and combines their results to achieve better performance than any individual model \cite{bolon2019ensembles, dong2020survey}.

Considering this, this paper combines a wrapper and an ensemble technique with the proposed NFISiS to enhance the models' ability to handle large datasets, optimize their performance, and increase interpretability, as only the most important features will be selected. The wrapper technique uses a GA to search for the combination of attributes that minimizes errors. These models are referred to as genetic NMR (GEN-NMR), genetic NTSK-RLS (GEN-NTSK-RLS), and genetic NTSK-wRLS (GEN-NTSK-wRLS). In addition, the Random NMR (R-NMR), Random NTSK (R-NTSK), and Random Forest NTSK (RF-NTSK) are introduced, all of which are ensemble models. The R-NMR and R-NTSK are ensemble versions of the NMR and NTSK models, respectively. On the other hand, RF-NTSK combines the outputs of R-NTSK and RF to compute the final result. The models are applied to renewable energy datasets. Such a series was chosen because of their complexity and relevance in many real-world applications. The models are applied to renewable energy datasets. These datasets were chosen due to their complexity and relevance in various real-world applications. 

Photovoltaic (PV) energy is expected to become one of the primary sources of energy worldwide in the future, as it is abundant, affordable, and easily scalable \cite{alcaniz2023trends}. However, the energy supplied by PV modules can be intermittent due to the varying nature of weather conditions \cite{notton2018intermittent}. Forecasting the power generated in a PV plant is crucial, as it helps grid operators, plant managers, and energy markets anticipate and manage the variability inherent in PV energy generation. Accurate forecasts optimize energy production, ensure a stable and reliable power supply, and contribute to cost reduction, efficient resource planning, and the overall sustainability of the energy landscape. Furthermore, the chaotic nature and uncertainties associated with PV energy make it difficult to obtain accurate results using physical equations, which has led to the increased use of machine learning models for predicting PV energy generation \cite{mayer2021extensive}.

Given the critical importance of PV energy control, numerous researchers have addressed this topic extensively. Wan et al. \cite{wan2015photovoltaic}, Barbieri et al. \cite{barbieri2017very}, Das et al. \cite{das2018forecasting}, Alcañiz et al. \cite{alcaniz2023trends} presented an extensive literature review on this topic. Munsif et al. \cite{munsif2023ct} proposed the convolutional-transformer-based network (CT-NET) for power forecasting. Jailani et al. \cite{jailani2023investigating} implement LSTM for solar energy. Artificial neural networks (ANN) \cite{khandakar2019machine}, ANN with feature selection \cite{o2017feature} was also discussed. Sharma et al. \cite{sharma2023solar} utilized a Levenberg Marquardt artificial neural network (LM-ANN) that uses the gradient descent (GD) optimization technique for solar energy forecasting. Among the advantages of using ANNs, the following can be highlighted: (i) self-adaptive ability; (ii) fault-tolerance; (iii) robustness; and (iv) strong inference capabilities \cite{yang2014weather}. However, due to the architecture of ANNs, they can present increased complexity. Support Vector Machines (SVM) have also been explored for PV energy forecasting \cite{preda2018pv}, praised for their non-linear modeling capacity and independence from prior knowledge \cite{li2016forecasting}. Numerous hybrid approaches using SVM have emerged, such as PSO-SVM \cite{eseye2018short}, GASVM \cite{vandeventer2019short}, SVM with an improved ant colony optimization (ACO-SVM). TSK-based models have some applications to the power predictions of PV systems \cite{liu2017takagi, liu2021novel}. A remarkable advantage of fuzzy models is their ability to handle uncertainties in real-world data \cite{ghofrani2016novel}. Still discussing fuzzy models, ANFIS usually presents errors substantially lower than those of fuzzy models. However, the computational complexity also increases, especially compared to SVM \cite{das2018forecasting, law2014direct}.

The remainder of this paper is organized as follows: Section \ref{NTSK} presents in detail the background knowledge on NTSK. Section \ref{proposals} introduces and discusses the proposed models, presenting the mathematical formalism. Section \ref{results} shows the simulations and discusses the results. Finally, Section \ref{conclusions} concludes de paper and proposes future works.

\section{Background Knowledge on NTSK}\label{NTSK}

The steps to define the NTSK rules are as follows:

\begin{itemize}
    \item \textbf{Step 1 - Define the Number of Rules:} Specify the maximum number of rules ($R_{\text{max}}$) as a hyper-parameter.
    
    \item \textbf{Step 2 - Compute Target Value Variations:} Calculate the variation in the target values for all samples.
    
    \item \textbf{Step 3 - Define Intervals for Variations:} Create equally spaced intervals based on the range of target value variations and $R_{\text{max}}$.
    
    \item \textbf{Step 4 - Assign Samples to Rules:} Assign samples to the corresponding rule based on their variation intervals.
    
    \item \textbf{Step 5 - Compute Fuzzy Sets:} Determine the parameters of the fuzzy sets (e.g., the mean and standard deviation for Gaussian membership functions).
    
    \item \textbf{Step 6 - Compute Consequent Parameters:} Estimate the consequent parameters using an adaptive filtering approach.
\end{itemize}

\begin{figure}[h]
    \centering
    \includegraphics[scale=0.6]{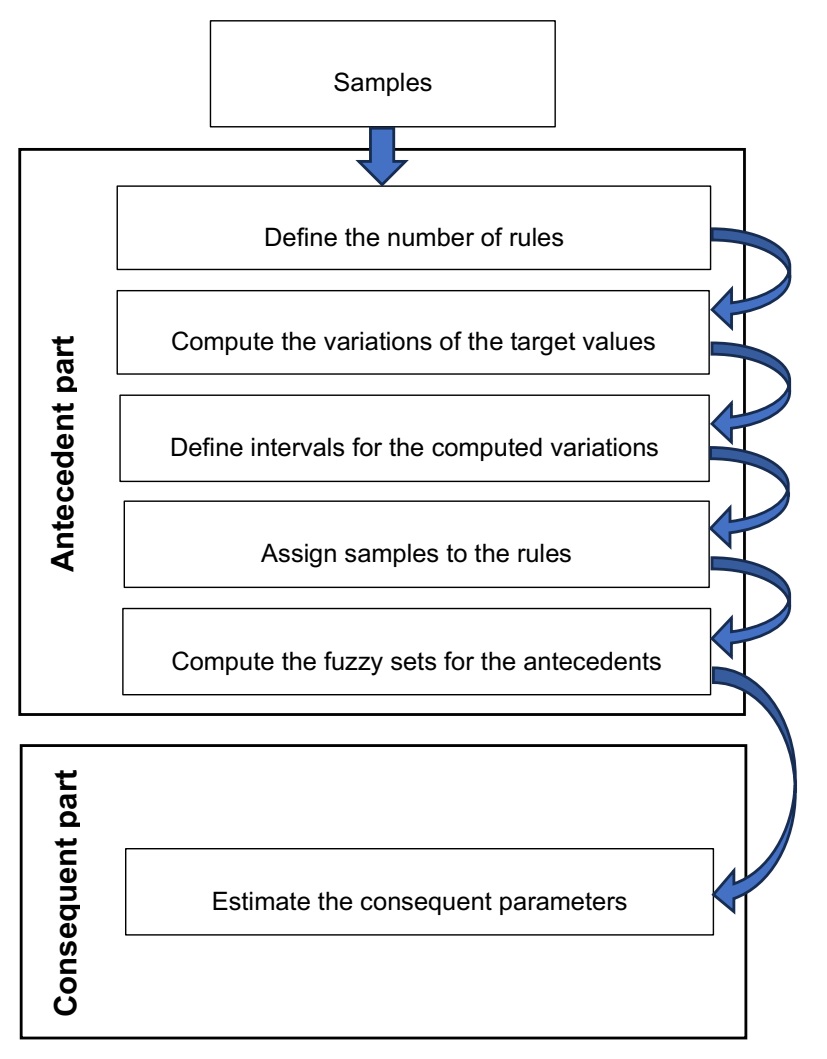}
    \caption{Flowchart presenting the learning phase for NTSK}
    \label{NTSK_}
\end{figure}

The flowchart for the NTSK training phase is presented in Figure \ref{NTSK_}.

\section{The Proposed Models}\label{proposals}

This section introduces the proposed models. Initially, the NMR is described and analyzed. Subsequently, the integration of GA and ensemble methods with NMR is presented and discussed.

\subsection{The New Mamdani Regressor (NMR)}

NMR is a data-driven fuzzy logic approach that constructs rules by defining intervals based on the target values. During the training phase, the model creates equally spaced intervals for the target variable to define the rules, subsequently determining the fuzzy sets for the antecedent part. The process is summarized as follows:

\textbf{Step 1 - Define the Number of Rules:} The user specifies the desired number of rules.

\textbf{Step 2 - Define Intervals for Target Values:} The model computes the amplitude of the target values and divides the result by the number of rules to create equally spaced intervals. 

\textbf{Step 3 - Define the Rules:} Each interval corresponds to one rule.

\textbf{Step 4 - Assign Samples to Rules:} Samples are assigned to the rule where the target value falls within the interval. 

\textbf{Step 5 - Compute the Fuzzy Sets:} Fuzzy sets are generated for the antecedent and consequent parts. 

\begin{figure}[htb]
    \centering
    \includegraphics[scale=0.6]{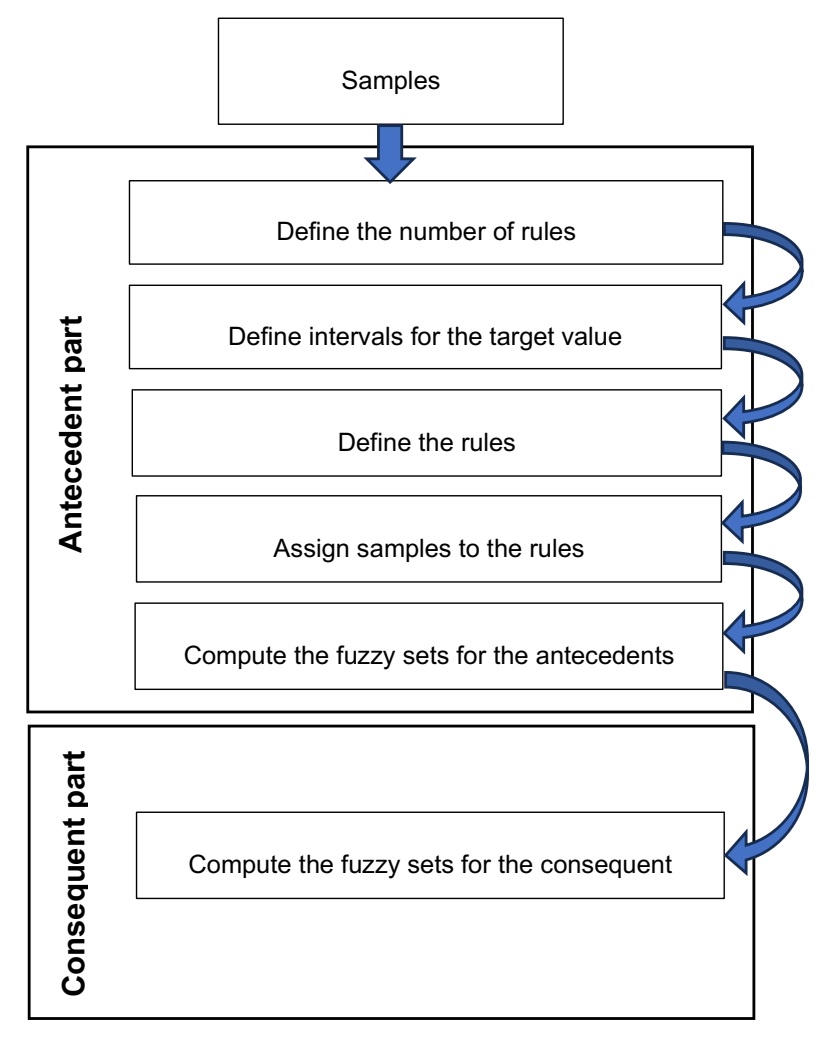}
    \caption{Flowchart of the learning phase for NMR}
    \label{FlowchartNMRTrain}
\end{figure}

Figure \ref{FlowchartNMRTrain} illustrates the steps in the training phase of NMR. Once trained, the model is ready for inference.

\subsubsection{Defining NMR Rules - Training Phase}

This part presents the mathematics and formalities of defining the rules. First, the user informs the desired number of rules, referred to as $R_{max}$. After that, the algorithm computes the interval size (IS) as a function of the target values' amplitude and $R_{max}$, as presented below:

\begin{equation}\label{IS}
    IS =  \frac{ \bar{y} - \underaccent{\bar}{y} }{R_{max}}
\end{equation}
where $IS$ means the range interval for the target value, $\bar{y}$ is the target's maximum value, $\underaccent{\bar}{y}$ is the target's minimum value, and $R_{max}$ is a hyper-parameter that defines the model's number of rules.

The $IS$ is used to compute each interval for the consequent part of all rules, as shown in Equation (\ref{Range}). The model assigns the samples to the rules based on the target value. So, for example, if the interval of the consequent part for the first rule is $[1,3]$, it means that the first rule comprises samples with a target value between one and three.

\begin{equation}\label{Range}
    Range_{i} = [\underaccent{\bar}{y} + (i - 1) IS,\underaccent{\bar}{y} + (i) IS]
\end{equation}
where $Range_{i}$ represents the interval of the consequent part of the $i$th rule and $i = 1, 2, \ldots, R_{max}$ is the rule number.

Defined the intervals of the consequent part of each rule, the next step consists of assigning the samples to their respective rule by comparing the target value with the intervals. Equation (\ref{i_NMR}) presents a formula to identify which rule the $k$th sample will be assigned.

\begin{equation}\label{i_NMR}
    i_{S^{k}} = \begin{cases}
        \lfloor \frac{ y^{k} - \underaccent{\bar}{y}  }{IS} \rfloor, & \mbox{if} \ \ y^{k} < \bar{y} \\
        R_{max}, & \mbox{otherwise}
    \end{cases}
\end{equation}
where $i_{S^{k}}$ is the rule in which the $k$th sample will be included and $\lfloor \rfloor$ represents the floor function.

Finally, the last step of defining the rules is to compute the fuzzy sets for the antecedent and consequent parts. Gaussian fuzzy sets are implemented in this paper (see Equation (\ref{Gaussian})). Only two parameters are required for Gaussian fuzzy sets: the mean and the standard deviation.

\begin{equation}\label{Gaussian}
    \mu_{A}(x) = \exp \left[ - \frac{1}{2} \left( \frac{x - v}{\sigma} \right)^{2} \right]
\end{equation}
where $v$ represents the mean of the fuzzy set and $\sigma$ is the standard deviation. Gaussian membership functions are advantageous because they are smooth, non-zero over all points, and effectively represent linguistic variables with precision and clarity \cite{azimi2020designing}.

\subsubsection{The Inference Process for NMR - Test Phase}

First, the input data is fuzzified using the respective membership functions. Then, each fuzzified input must be combined using a fuzzy operator to compute the normalized firing degree for each rule (Equation (\ref{Firing_norm})). The rule with a higher degree of activation informs a higher degree of pertinence with the respective input. 

\begin{equation}\label{Firing_norm}
    w_{i} = \frac{ \prod_{j=1}^{p} \exp{ \left[ - \frac{1}{2} \frac{\left( x_{j}^{k} - v_{i,j} \right)^{2}}{\sigma_{i,j}^{2}} \right]} }{ \sum_{i=1}^{R} \left( \prod_{j=1}^{p} \exp{ \left[ - \frac{1}{2} \frac{\left( x_{j}^{k} - v_{i,j} \right)^{2}}{\sigma_{i,j}^{2}} \right]} \right) }
\end{equation}
where $v_{i,j}$ is mean of the $j$th attribute for the $i$th rule, $v_{i} = [v_{i,1}, \ldots, v_{i,p}]^{T}$, $p$ is the dimension of the inputs (number of attributes for each sample), $x_{j}^{k}$ is the $k$th input vector for the $j$th attribute, and $\sigma_{i,j}$ is the standard deviation of the $j$th attribute for the $i$th rule, for $\sigma_{i} = [\sigma_{i,1}, \ldots, \sigma_{i,p}]^{T}$.

Finally, defuzzification can occur, which consists of computing the model's output according to Equation (\ref{Output}).

\begin{equation}\label{Output}
    \hat{y}^{k} = \frac{\sum_{i=1}^{R_{max}} v_{i,output} \times w_{i}}{\sum_{i=1}^{R_{max}} w_{i}}
\end{equation}
where $v_{i,output}$ is mean of the consequent fuzzy set for the $i$th rule.

The concepts of NMR can easily be expanded to the new Mamdani classifier (NMC), but instead of having the number of rules defined by the user, it will be related to the number of classes the data have. Each class will constitute a rule. However, it will not be discussed here.

\subsubsection{Genetic Algorithm Implementation}

This section presents the implementation of the Genetic Algorithm (GA) for feature selection. Initially, the dataset is divided into training and testing subsets. A binary vector, whose length corresponds to the number of attributes, is then created to represent feature selection. Each position in the vector indicates whether the corresponding attribute is included (1) or excluded (0) in the feature subset. To ensure at least one attribute is always selected, the vector must contain at least one nonzero element; otherwise, no attributes would be chosen.

Once the initial population of vectors is generated, the algorithm evaluates their fitness by invoking the fuzzy model. The fitness assessment is based on the model's performance using the selected features. The GA iteratively evolves the population until the termination criteria are met. At the end of the process, the vector yielding the lowest error is saved as the optimal feature subset.

\subsection{Ensemble Fuzzy}

This section introduces three ensemble fuzzy models: R-NMR, R-NTSK, and RF-NTSK. The R-NMR and R-NTSK models operate similarly, employing a randomized approach to attribute selection. In both models, a subset of attributes is randomly selected from the original dataset, and $z$ models are trained. The best-performing model from this subset is added to the ensemble, after which the process is repeated to identify additional models until the desired number of inducers is obtained.

Finally, RF-NTSK acts as an ensemble of ensembles, combining the outputs of RF and NTSK. The final prediction is calculated using Equation (\ref{RF-NTSK}), which assigns weights to each output based on their respective errors.

\begin{equation}\label{RF-NTSK}
    \hat{y}^{k} = \hat{y}_{RF}^{k} \frac{\epsilon_{R-NTSK}}{\epsilon_{RF} + \epsilon_{R-NTSK}} + \hat{y}_{R-NTSK}^{k} \frac{\epsilon_{RF}}{\epsilon_{RF} + \epsilon_{R-NTSK}}
\end{equation}
where $\epsilon_{RF}$ and $\epsilon_{R-NTSK}$ are the estimated error of RF and R-NTSK during the training phase, respectively, and $\hat{y}_{RF}^{k}$ and $\hat{y}_{R-NTSK}^{k}$ the output of RF and R-NTSK, respectively. It can be seen that the output is a weighted average of the output of RF and R-NTSK where the weights are inversely proportional to the errors.

\section{Experimental Results and Discussion}\label{results}

This section presents the metrics employed to evaluate the simulations and shows and discusses the results.

\subsection{Evaluation Indicators}

The model's performance is assessed using the normalized root-mean-square error (NRMSE), the non-dimensional index error (NDEI), and the mean absolute percentage error (MAPE), defined by Equations (\ref{nrmse}), (\ref{ndei}), and (\ref{mape}), respectively.

\begin{equation}\label{nrmse}
    NRMSE = \frac{RMSE}{\bar{y} - \underaccent{\bar}{y}}
\end{equation}

\begin{equation}\label{ndei}
    NDEI = \frac{RMSE}{std([y^{1},...,y^{T}])}
\end{equation}

\begin{equation}\label{mape}
    MAPE = \frac{1}{T} \sum_{k=1}^{T} \frac{\| y^{k} - \hat{y}^{k} \| }{y^{k}}
\end{equation}
where $y^{k}$ is the $k$th actual value, $\hat{y}^{k}$ is the $k$th predicted value, $\bar{y}$ is the maximum value for $y$, $\underaccent{\bar}{y}$ is the minimum value for $y$, $T$ is the sample size, and $std()$ is the standard deviation function, and RMSE is the root-mean-square error given in Equation (\ref{rmse}). 

\begin{equation}\label{rmse}
    RMSE = \sqrt{\frac{1}{T}\sum_{k=1}^{T}(y^{k}-\hat{y}^{k})^{2}}
\end{equation}

RMSE is largely used in the literature due to its advantages, such as being sensitive to outliers, suitable for optimizations, and allowing comparison between different models. To improve the visualization, the paper reports the normalized NRMSE, which normalizes the RMSE by the amplitude. Furthermore, the NDEI is commonly used in the literature to evaluate eFS models. This metric informs how many times the RMSE is concerning the standard deviation of the data. On the other hand, the MAPE is less sensitive to outliers, but it is suitable in cases where the percentage error is more relevant than absolute errors.

For rule-based models, the total number of final rules is also reported. The hyper-parameters of the models are optimized through a grid search to achieve the lowest possible error.

Additionally, two statistical tests are employed to analyze the linearity and stationarity of real-world time series: the Shapiro-Wilk test \cite{shapiro1965analysis} and the augmented Dickey-Fuller (ADF) test \cite{dickey1981likelihood}. While the derivations of these tests are beyond the scope of this work, detailed explanations can be found in the literature.

For the Shapiro-Wilk test, a $p$-value less than 0.05 rejects the null hypothesis, indicating that the series does not originate from a normal distribution. Conversely, a $p$-value greater than 0.05 suggests that the series follows a normal distribution. Similarly, the ADF test interprets a $p$-value below 0.05 as evidence supporting stationarity, while a $p$-value exceeding 0.05 implies non-stationarity.

\subsection{Datasets}

Finally, PV energy datasets from two power plants, Alice Springs and Yulara Solar System, are implemented. Desert Knowledge Australia Solar Centre (DKASC) Alice Springs is a giant solar energy power station located in the Alice Springs desert that contains PV technologies of different types, ages, makes, models, and configurations. Operating since 2008, Alice offers a vast database for researchers. This work aims to predict the daily power one step ahead using as predictors the following attributes: humidity, diffuse radiation, radiation, diffuse tilted, accumulated energy, rainfall, wind direction, temperature, radiation global, tilted, energy, global radiation, power, and the current. 

On the other hand, the Yulara Solar System, installed in 2014, is located near Yulara and operates a 1.8 MW solar photovoltaic plant. Two datasets from Yulara are implemented to predict the daily power one step ahead using as predictors air pressure, wind direction, rainfall, hail, wind speed, accumulated energy, max wind speed, pyranometer, global radiation, temperature, temperature probe 1, temperature probe 2, energy, power, and the current. More information about these attributes can be found on the website. The PV panels Yulara 1 and 5 were chosen to perform the simulations. The initial dataset contained information in the 5-minute interval. Pre-processing was implemented to remove null values and convert the information to daily values. The datasets' period includes data from January 2021 to December 2022.

More information about the datasets can be found on the official website: \url{https://dkasolarcentre.com.au/}. As the plant uses solar panels with different characteristics, the panels are identified by numbers. Table \ref{SolarPanelCharacteristics} presents the characteristics of PV panels, and Table \ref{datasetstats} shows the statistical results for the datasets. The results suggest that only Yulara 5 comes from a normal distribution and that all datasets are stationary.

% Table generated by Excel2LaTeX
\begin{table}[htbp]
    \caption{Characteristics of the PV panels implemented in the simulations}
    \begin{center}
    \scalebox{1}{
    \begin{tabular}{ccccc}
    \hline
     Characteristics & Alice 1A & Alice 38 & Yulara 1 & Yulara 5 \\
    \hline
    Manufacturer & Trina & Q CELLS & - & -  \\
    Array Rating & 10.5kW & 5.9kW & 1058.4kW & 105.9kW \\
    PV Technology & mono-Si & mono-Si & poly-Si & mono-Si \\
    Array Structure	Tracker & Dual Axis & Ground Mount & Ground Mount & Roof Mount \\
    Installed & 2009 & 2017 & 2016 & 2016 \\
    \hline
    \end{tabular}}%
    \end{center}
    \label{SolarPanelCharacteristics}%
\end{table}%

% Table generated by Excel2LaTeX
\begin{table}[htbp]
    \caption{Statistical tests of the PV energy datasets}
    \begin{center}
    \scalebox{1}{
    \begin{tabular}{ccccc}
    \hline
     Test & Alice 1A & Alice 38 & Yulara 1 & Yulara 5 \\
    \hline
    Shapiro-Wilk & $5.0 \times 10^{-17}$ & $1.4 \times 10^{-27}$ & $5.4 \times 10^{-11}$ & 0.37 \\
    ADF & $8.5 \times 10^{-8}$ & $1.0 \times 10^{-20}$ & $3.7 \times 10^{-20}$ & $8.3 \times 10^{-4}$ \\
    \hline
    \end{tabular}}%
    \end{center}
    \label{datasetstats}%
\end{table}%

\subsection{Results}

\begin{figure}
    \centering
    \includegraphics[scale=0.3]{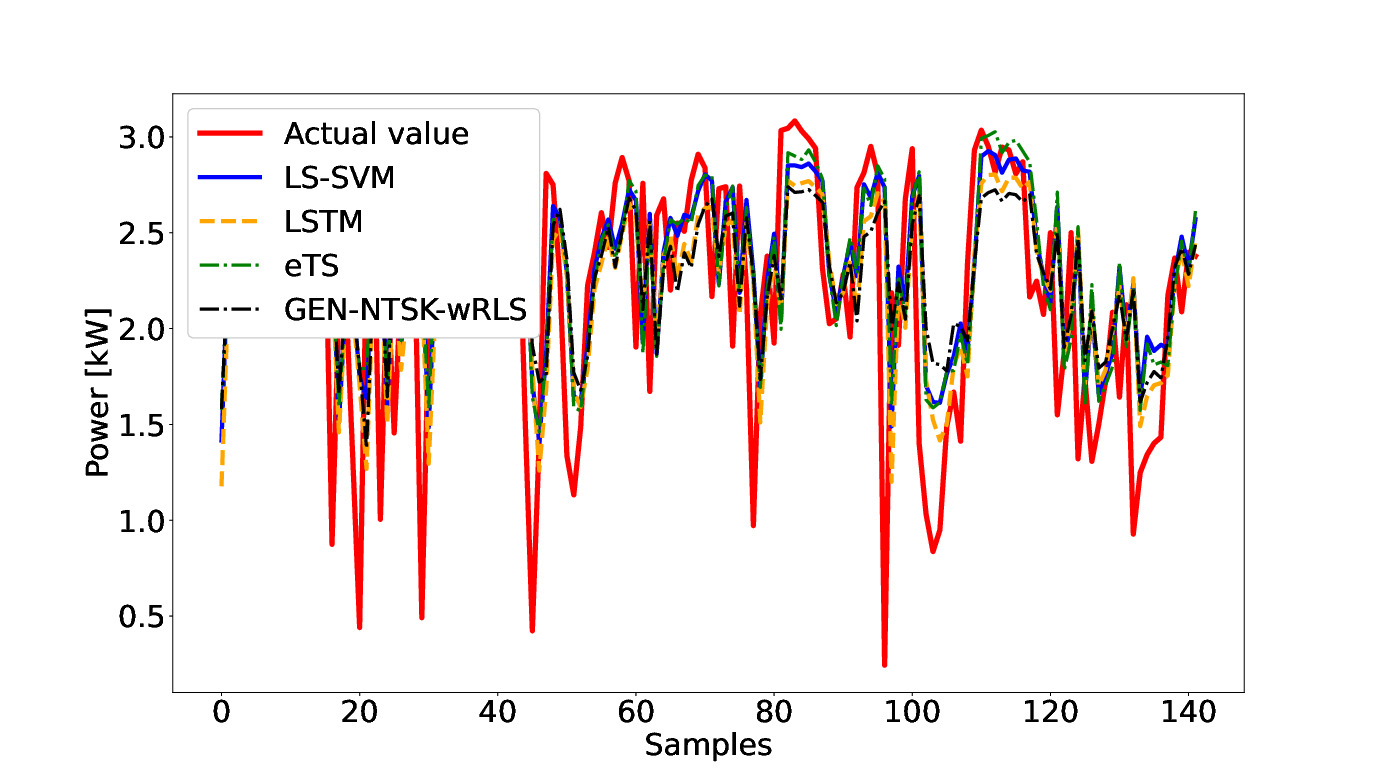}
    \caption{Graphic of predictions for Alice 1A}
    \label{GraphicAlice1A}
\end{figure}

Table~\ref{TabAlice1A} presents the simulations' results for the Alice 1A time series. SVM and LS-SVM obtained the lowest errors among the classical models, LSTM the lowest among DL models, eTS and ePL+ the lowest for the eFSs, and GEN-NTSK (wRLS) and R-NTSK for the proposed fuzzy models. Furthermore, eTS obtained the lowest NRMSE and NDEI among all simulations and ePL+ the lowest MAPE, supporting the good performance of the eFSs in real datasets. The ePL, NTSK (RLS), and GEN-NTSK (RLS) achieved the fewest rules and eMG the highest ones. Figure~\ref{GraphicAlice1A} presents the predictions of the lowest NRMSE for each model class for the Alice 1A time series. The models failed to predict some extreme points, which may indicate that the presence of outliers disturbs the predictions.

% Table generated by Excel2LaTeX
\begin{table}[htbp]
    \caption{Simulations' results for the Alice 1A PV panel}
    \begin{center}
    \scalebox{0.9}{
    \begin{tabular}{ccccc}
    \hline
     Model & NRMSE & NDEI & MAPE & Rules \\
    \hline
    KNN \cite{fix1951discriminatory} & 0.22656 & 1.00709 & 0.40166 & - \\
    Regression Tree \cite{breiman1984classification} & 0.21325 & 0.94790 & 0.37171 & - \\
    Random Forest \cite{ho1995random} & 0.21480 & 0.95481 & 0.36574 & - \\
    SVM \cite{cortes1995support} & 0.20714 & 0.92075 & \textbf{0.35383} & - \\
    LS-SVM \cite{suykens1999least} & \textbf{0.20669} & \textbf{0.91876} & 0.35509 & - \\
    GBM \cite{friedman2001greedy} & 0.21849 & 0.97119 & 0.36887 & - \\
    XGBoost \cite{chen2016xgboost} & 0.21538 & 0.95736 & 0.37573 & - \\
    LGBM \cite{ke2017lightgbm} & 0.21320 & 0.94771 & 0.37708 & - \\
    \hline
    MLP \cite{rosenblatt1958perceptron} & 0.26037 & 1.15734 & 0.46606 & - \\
    CNN \cite{fukushima1980neocognitron} & 0.22639 & 1.00634 & 0.40184 & - \\
    RNN \cite{hopfield1982neural} & 0.20881 & 0.92818 & 0.36124 & - \\
    LSTM \cite{hochreiter1997long} & \textbf{0.20776} & \textbf{0.92350} & \textbf{0.35657} & - \\
    GRU \cite{chung2014empirical} & 0.26014 & 1.15636 & 0.39765 & - \\
    WaveNet \cite{oord2016wavenet} & 0.25401 & 1.12910 & 0.45572 & - \\
    \hline
    eTS \cite{angelov2004approach} & \textbf{0.20639} & \textbf{0.91742} & 0.35508 & 4 \\
    Simpl\_eTS \cite{angelov2005simpl_ets} & 0.33692 & 1.49764 & 0.43992 & 56 \\
    exTS \cite{angelov2006evolving} & 0.20861 & 0.92729 & 0.36044 & 3 \\
    ePL \cite{lima2010evolving} & 0.20953 & 0.93136 & 0.35760 & 1 \\
    eMG \cite{lemos2010multivariable} & 0.40435 & 1.79734 & 0.56272 & 140 \\
    ePL+ \cite{maciel2012enhanced} & 0.20757 & 0.92266 & \textbf{0.34947} & 3 \\
    ePL-KRLS-DISCO \cite{alves2021novel} & 0.29263 & 1.30077 & 0.46094 & 33 \\
    \hline
    NMR & 0.24337 & 1.08178 & 0.38902 & 16 \\
    NTSK (RLS) & 0.24953 & 1.10918 & 0.43855 & 1 \\
    NTSK (wRLS) & 0.20957 & 0.93154 & 0.35129 & 4 \\
    GEN-NMR & 0.23421 & 1.04108 & 0.38867 & 17 \\
    GEN-NTSK (RLS) & 0.22072 & 0.98112 & 0.36194 & 1 \\
    GEN-NTSK (wRLS) & \textbf{0.20711} & \textbf{0.92061} & 0.36308 & 19 \\
    R-NMR & 0.23220 & 1.03216 & 0.36129 & - \\
    R-NTSK & 0.21431 & 0.95261 & \textbf{0.35784} & - \\
    RF-NTSK & 0.21177 & 0.94132 & 0.35908 & - \\
    \hline
    \end{tabular}}%
    \end{center}
    \label{TabAlice1A}%
\end{table}%

\begin{figure}
    \centering
    \includegraphics[scale=0.3]{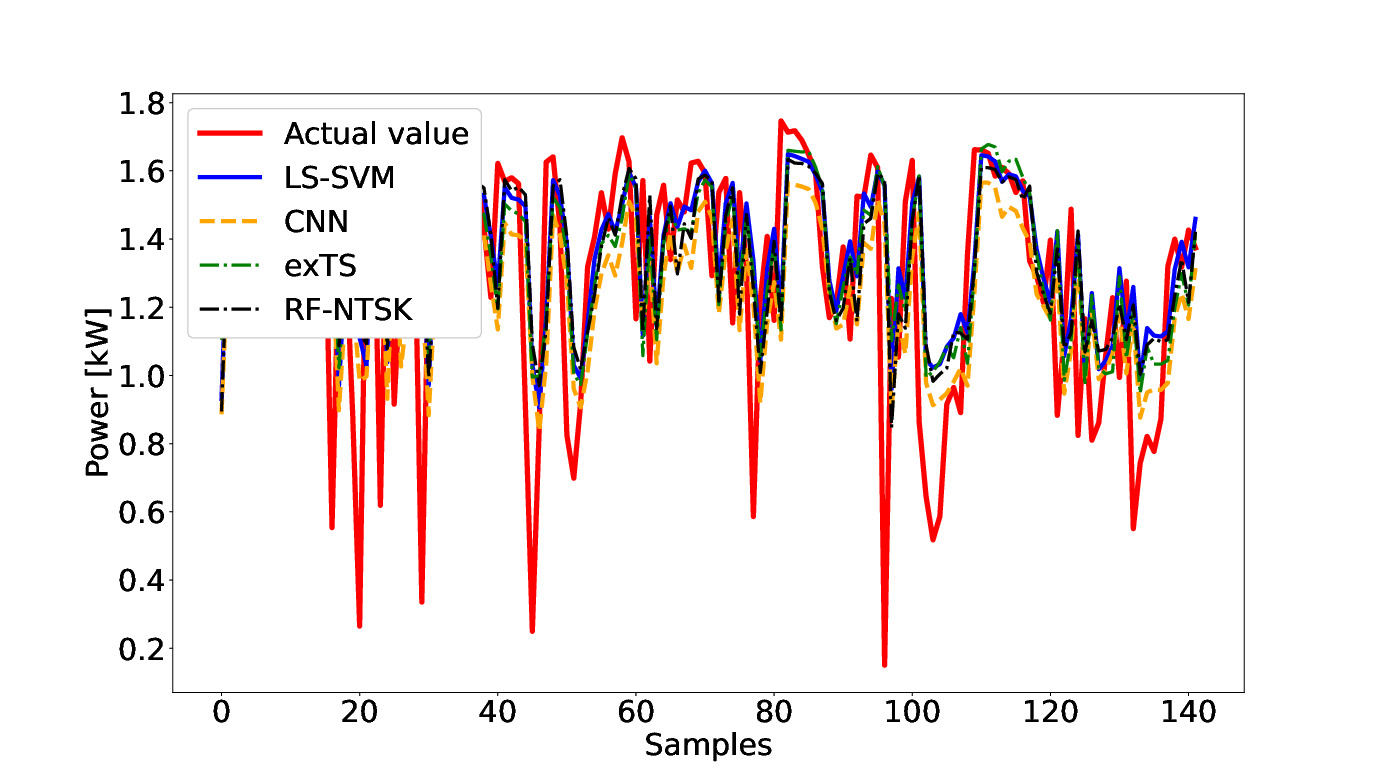}
    \caption{Graphic of predictions for Alice 38}
    \label{GraphicAlice38}
\end{figure}

Table~\ref{TabAlice38} presents the simulations' results for the Alice 38 time series. LS-SVM obtained the lowest errors among the classical models, CNN and GRU the lowest among DL models, eTS and exTS the lowest for the eFSs, and R-NTSK and RF-NTSK for the proposed fuzzy models. The ePL, NTSK (RLS), and GEN-NTSK (RLS) achieved the fewest final rules, and eMG the highest ones. Figure~\ref{GraphicAlice38} presents the best predictions of each class for the Alice 38 time series.

% Table generated by Excel2LaTeX
\begin{table}[htbp]
    \caption{Simulations' results for the Alice 38 PV panel}
    \begin{center}
    \scalebox{0.9}{
    \begin{tabular}{ccccc}
    \hline
     Model & NRMSE & NDEI & MAPE & Rules \\
    \hline
    KNN \cite{fix1951discriminatory} & 0.23499 & 1.04722 & 0.37156 & - \\
    Regression Tree \cite{breiman1984classification} & 0.21404 & 0.95385 & 0.34513 & - \\
    Random Forest \cite{ho1995random} & 0.21363 & 0.95204 & 0.34780 & - \\
    SVM \cite{cortes1995support} & 0.21480 & 0.95723 & 0.34494 & - \\
    LS-SVM \cite{suykens1999least} & \textbf{0.20813} & \textbf{0.92755} & \textbf{0.33040} & - \\
    GBM \cite{friedman2001greedy} & 0.25371 & 1.13065 & 0.37142 & - \\
    XGBoost \cite{chen2016xgboost} & 0.22014 & 0.98103 & 0.34765 & - \\
    LGBM \cite{ke2017lightgbm} & 0.21548 & 0.96026 & 0.34166 & - \\
    \hline
    MLP \cite{rosenblatt1958perceptron} & 0.21231 & 0.94616 & 0.34199 & - \\
    CNN \cite{fukushima1980neocognitron} & \textbf{0.20700} & \textbf{0.92250} & 0.32452 & - \\
    RNN \cite{hopfield1982neural} & 0.21020 & 0.93677 & 0.33125 & - \\
    LSTM \cite{hochreiter1997long} & 0.20858 & 0.92953 & 0.33252 & - \\
    GRU \cite{chung2014empirical} & 0.21089 & 0.93981 & \textbf{0.32443} & - \\
    WaveNet \cite{oord2016wavenet} & 0.25446 & 1.13400 & 0.42215 & - \\
    \hline
    eTS \cite{angelov2004approach} & 0.21114 & 0.94093 & \textbf{0.32321} & 3 \\
    Simpl\_eTS \cite{angelov2005simpl_ets} & 0.22844 & 1.01805 & 0.34967 & 36 \\
    exTS \cite{angelov2006evolving} & \textbf{0.20443} & \textbf{0.91105} & 0.33023 & 3 \\
    ePL \cite{lima2010evolving} & 0.58321 & 2.59907 & 0.68757 & 1 \\
    eMG \cite{lemos2010multivariable} & 0.28917 & 1.28866 & 0.42192 & 138 \\
    ePL+ \cite{maciel2012enhanced} & 0.25132 & 1.11999 & 0.35191 & 11 \\
    ePL-KRLS-DISCO \cite{alves2021novel} & 0.26151 & 1.16542 & 0.36657 & 33 \\
    \hline
    NMR & 0.24519 & 1.09269 & 0.39498 & 9 \\
    NTSK (RLS) & 0.25395 & 1.13171 & 0.41610 & 1 \\
    NTSK (wRLS) & 0.23656 & 1.05424 & 0.35931 & 11 \\
    GEN-NMR & 0.26173 & 1.16638 & 0.35348 & 13 \\
    GEN-NTSK (RLS) & 0.21581 & 0.96176 & 0.34244 & 1 \\
    GEN-NTSK (wRLS) & 0.25224 & 1.12411 & 0.36122 & 13 \\
    R-NMR & 0.21897 & 0.97582 & 0.33655 & - \\
    R-NTSK & 0.21150 & 0.94252 & \textbf{0.33032} & - \\
    RF-NTSK & \textbf{0.21087} & \textbf{0.93974} & 0.33716 & - \\
    \hline
    \end{tabular}}%
    \end{center}
    \label{TabAlice38}%
\end{table}%

%\subsubsection{Yulara Solar System}

% Table generated by Excel2LaTeX

\begin{figure}
    \centering
    \includegraphics[scale=0.3]{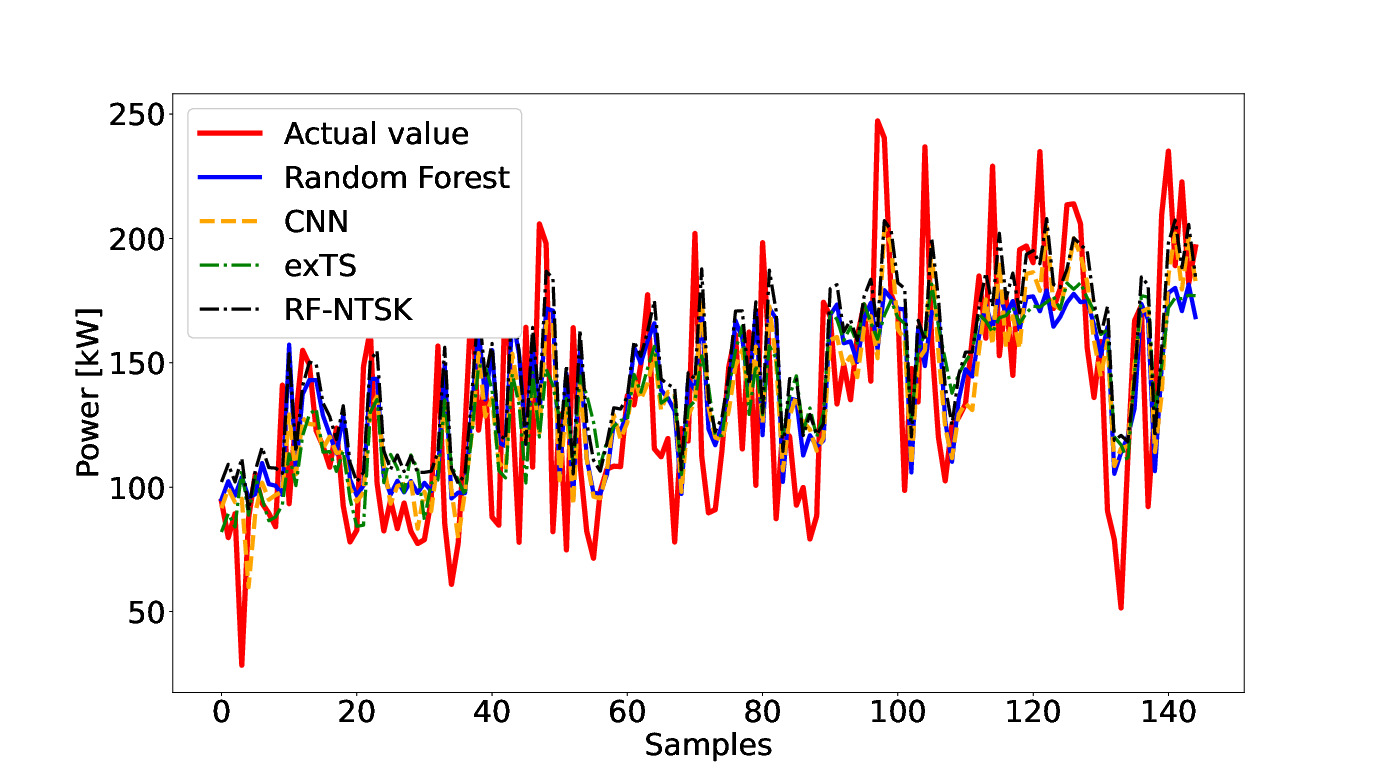}
    \caption{Graphic of predictions for Yulara 1}
    \label{GraphicYulara1}
\end{figure}

Table~\ref{TabYulara1} presents the simulations' results for the Yulara 1 time series. RF obtained the lowest errors among the classical models, CNN among the DL models, eTS and exTS the lowest for the eFSs, and GEN-NMR and RF-NTSK for the proposed fuzzy models. CNN obtained the lowest among all models. The eMG model achieved 189 final rules, the highest among all models. The ePL, ePL+, NTSK (RLS), and GEN-NTSK (RLS) obtained just one final rule. Figure~\ref{GraphicYulara1} presents the best predictions of each class for the Yulara 1 time series.

% Table generated by Excel2LaTeX
\begin{table}[htbp]
    \caption{Simulations' results for the Yulara 1 PV panel}
    \begin{center}
    \scalebox{0.9}{
    \begin{tabular}{ccccc}
    \hline
     Model & NRMSE & NDEI & MAPE & Rules \\
    \hline
    KNN \cite{fix1951discriminatory} & 0.21399 & 1.02255 & 0.39364 & - \\
    Regression Tree \cite{breiman1984classification} & 0.21037 & 1.00523 & 0.28632 & - \\
    Random Forest \cite{ho1995random} & \textbf{0.17592} & \textbf{0.84064} & \textbf{0.27256} & - \\
    SVM \cite{cortes1995support} & 0.21055 & 1.00610 & 0.34657 & - \\
    LS-SVM \cite{suykens1999least} & 0.21319 & 1.01869 & 0.36743 & - \\
    GBM \cite{friedman2001greedy} & 0.18105 & 0.86514 & 0.29454 & - \\
    XGBoost \cite{chen2016xgboost} & 0.20440 & 0.97670 & 0.30765 & - \\
    LGBM \cite{ke2017lightgbm} & 0.17896 & 0.85515 & 0.28816 & - \\
    \hline
    MLP \cite{rosenblatt1958perceptron} & 0.17511 & 0.83676 & 0.26667 & - \\
    CNN \cite{fukushima1980neocognitron} & \textbf{0.16568} & \textbf{0.79168} & \textbf{0.25176} & - \\
    RNN \cite{hopfield1982neural} & 0.22641 & 1.08191 & 0.29806 & - \\
    LSTM \cite{hochreiter1997long} & 0.24267 & 1.15959 & 0.29554 & - \\
    GRU \cite{chung2014empirical} & 0.22945 & 1.09639 & 0.29685 & - \\
    WaveNet \cite{oord2016wavenet} & 0.17783 & 0.84973 & 0.27339 & - \\
    \hline
    eTS \cite{angelov2004approach} & 0.18750 & 0.89595 & \textbf{0.25613} & 4 \\
    Simpl\_eTS \cite{angelov2005simpl_ets} & 0.20622 & 0.98540 & 0.28118 & 45 \\
    exTS \cite{angelov2006evolving} & \textbf{0.16995} & \textbf{0.81211} & 0.25916 & 5 \\
    ePL \cite{lima2010evolving} & 0.22578 & 1.07888 & 0.39487 & 1 \\
    eMG \cite{lemos2010multivariable} & 0.28605 & 1.36686 & 0.51978 & 176 \\
    ePL+ \cite{maciel2012enhanced} & 0.21503 & 1.02751 & 0.36951 & 1 \\
    ePL-KRLS-DISCO \cite{alves2021novel} & 0.21213 & 1.01365 & 0.36978 & 20 \\
    \hline
    NMR & 0.27135 & 1.29663 & 0.38104 & 8 \\
    NTSK (RLS) & 0.20606 & 0.98463 & 0.34618 & 1 \\
    NTSK (wRLS) & 0.22261 & 1.06370 & 0.38723 & 2 \\
    GEN-NMR & 0.19319 & 0.92315 & \textbf{0.26033} & 19 \\
    GEN-NTSK (RLS) & 0.26105 & 1.24740 & 0.45110 & 1 \\
    GEN-NTSK (wRLS) & 0.21812 & 1.04229 & 0.37640 & 19 \\
    R-NMR & 0.26611 & 1.27157 & 0.36162 & - \\
    R-NTSK & 0.20941 & 1.00066 & 0.35864 & - \\
    RF-NTSK & \textbf{0.18351} &\textbf{ 0.87687} & 0.30299 & - \\
    \hline
    \end{tabular}}%
    \end{center}
    \label{TabYulara1}%
\end{table}%

\begin{figure}
    \centering
    \includegraphics[scale=0.3]{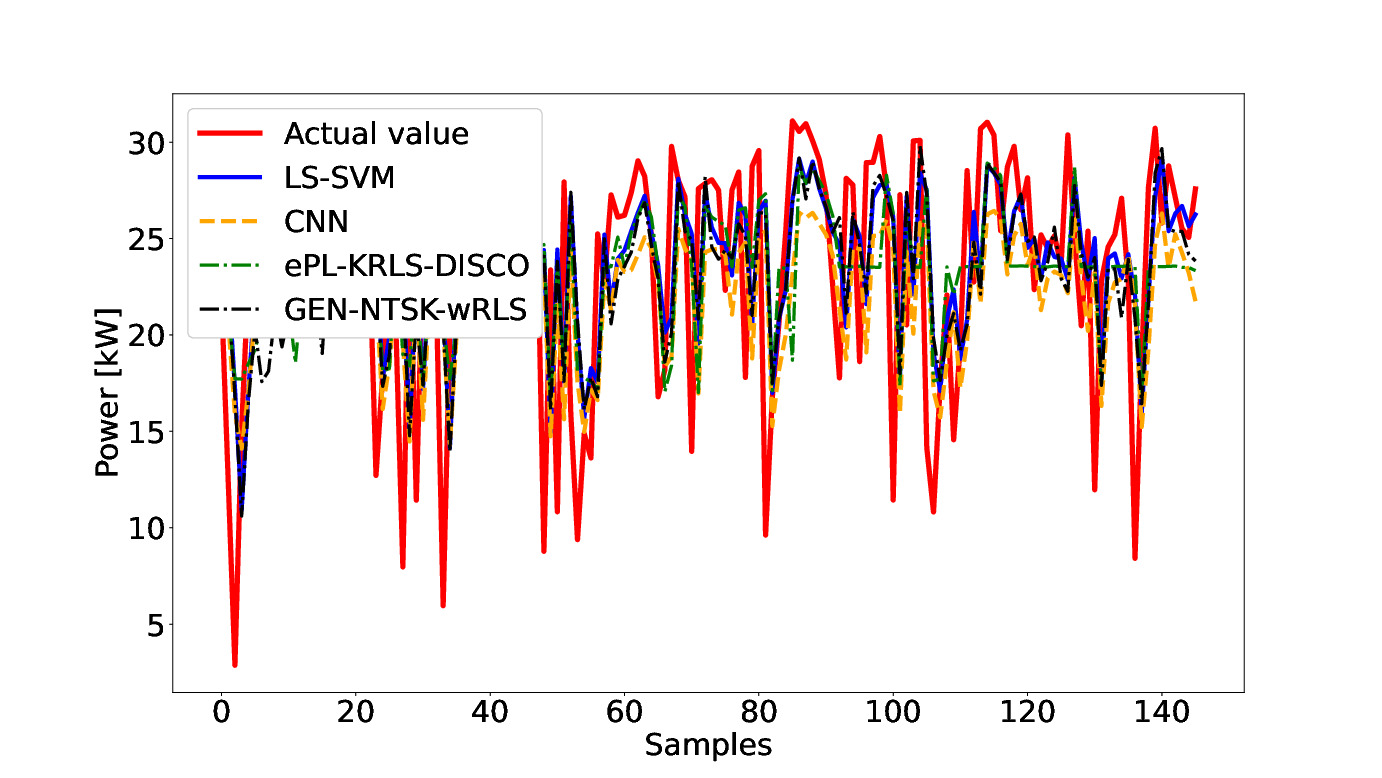}
    \caption{Graphic of predictions for Yulara 5}
    \label{GraphicYulara5}
\end{figure}

Table~\ref{TabYulara5} presents the simulations' results for the Yulara 5 time series. LS-SVM obtained the lowest errors among the classical models, CNN and GRU the lowest among the DL, ePL-KRLS-DISCO the lowest for the eFSs, and GEN-NTSK (wRLS) and R-NTSK the lowest errors for the proposed models. Among all models, ePL-KRLS-DISCO achieved the lowest NRMSE and NDEI, and R-NTSK the lowest MAPE. Simpl\_eTS performed 431 final rules, the highest among all models, and eMG had 372 final rules, the second-highest ones. Figure~\ref{GraphicYulara5} presents the best predictions of each class for the Yulara 5 time series.

% Table generated by Excel2LaTeX
\begin{table}[htbp]
    \caption{Simulations' results for the Yulara 5 PV panel}
    \begin{center}
    \scalebox{0.9}{
    \begin{tabular}{ccccc}
    \hline
     Model & NRMSE & NDEI & MAPE & Rules \\
    \hline
    KNN \cite{fix1951discriminatory} & 0.23408 & 1.10290 & 0.33711 & - \\
    Regression Tree \cite{breiman1984classification} & 0.21492 & 1.01262 & 0.29218 & - \\
    Random Forest \cite{ho1995random} & 0.21306 & 1.00388 & 0.28702 & - \\
    SVM \cite{cortes1995support} & 0.22354 & 1.05322 & 0.29446 & - \\
    LS-SVM \cite{suykens1999least} & \textbf{0.20783} & \textbf{0.97920} & \textbf{0.28330} & - \\
    GBM \cite{friedman2001greedy} & 0.21769 & 1.02570 & 0.28887 & - \\
    XGBoost \cite{chen2016xgboost} & 0.22848 & 1.07649 & 0.30705 & - \\
    LGBM \cite{ke2017lightgbm} & 0.21589 & 1.01717 & 0.28938 & - \\
    \hline
    MLP \cite{rosenblatt1958perceptron} & 0.21564 & 1.01601 & 0.30108 & - \\
    CNN \cite{fukushima1980neocognitron} & \textbf{0.21186} & \textbf{0.99821} & 0.28915 & - \\
    RNN \cite{hopfield1982neural} & 0.29585 & 1.39393 & 0.37098 & - \\
    LSTM \cite{hochreiter1997long} & 0.30403 & 1.43246 & 0.37513 & - \\
    GRU \cite{chung2014empirical} & 0.22070 & 1.03985 & \textbf{0.28804} & - \\
    WaveNet \cite{oord2016wavenet} & 0.21745 & 1.02455 & 0.29647 & - \\
    \hline
    eTS \cite{angelov2004approach} & 0.21658 & 1.02046 & 0.28506 & 3 \\
    Simpl\_eTS \cite{angelov2005simpl_ets} & 0.22494 & 1.05983 & 0.29851 & 431 \\
    exTS \cite{angelov2006evolving} & 0.20669 & 0.97386 & 0.28372 & 3 \\
    ePL \cite{lima2010evolving} & 0.20992 & 0.98907 & 0.29031 & 1 \\
    eMG \cite{lemos2010multivariable} & 0.32813 & 1.54603 & 0.41719 & 372 \\
    ePL+ \cite{maciel2012enhanced} & 0.20917 & 0.98555 & 0.28964 & 4 \\
    ePL-KRLS-DISCO \cite{alves2021novel} & \textbf{0.20501} & \textbf{0.96595} & \textbf{0.28255} & 20 \\
    \hline
    NMR & 0.32884 & 1.54935 & 0.39332 & 2 \\
    NTSK (RLS) & 0.22291 & 1.05029 & 0.31533 & 1 \\
    NTSK (wRLS) & 0.21398 & 1.00818 & 0.28825 & 5 \\
    GEN-NMR & 0.29698 & 1.39928 & 0.36355 & 15 \\
    GEN-NTSK (RLS) & 0.22080 & 1.04035 & 0.29857 & 1 \\
    GEN-NTSK (wRLS) & \textbf{0.20625} & \textbf{0.97176} & 0.28503 & 15 \\
    R-NMR & 0.36554 & 1.72230 & 0.42459 & - \\
    R-NTSK & 0.20632 & 0.97209 & \textbf{0.27714} & - \\
    RF-NTSK & 0.20749 & 0.97761 & 0.27978 & - \\
    \hline
    \end{tabular}}%
    \end{center}
    \label{TabYulara5}%
\end{table}%

\subsection{Rules and Interpretability}

A key advantage of genetic-based approaches is their ability to optimize the trade-off between accuracy and interpretability. For instance, the PV energy dataset contains numerous attributes. The GEN-NTSK-wRLS method selected six out of the twelve attributes, prioritizing those that most effectively contributed to predictive performance. This demonstrates one of the primary strengths of genetic algorithms: they can uncover hidden correlations between the target variable and the attributes, which may not be immediately evident through simple correlation analysis. As a result, the model identified a more concise rule-based structure with fewer rules, enhancing the analysis by focusing on the attributes that most significantly influence the outcomes.

% Table generated by Excel2LaTeX
\begin{table}[!h]
    \caption{Rules of GEN-NTSK-wRLS for Alice 1A with four rules}
    \begin{center}
    \scalebox{0.9}{
    \begin{tabular}{c|cccccc|c}
    \hline
     Rule & Diffuse Radiation & Rainfall & Wind Direction & Current & Power & Global Radiation & Next Power \\
    \hline
    1 & 57.67 (27.35) & 0.07 (0.20) & 24.22 (18.12) & 5.04 (0.85) & 2.48 (0.49) & 301.66 (69.41) & [-2.56, -1.37] \\
    2 & 61.40 (37.21) & 0.56 (2.41) & 31.20 (15.20) & 4.48 (1.07) & 2.18 (0.64) & 264.34 (88.11) & [-1.37, -0.18] \\   
    3 & 54.35 (35.91) & 0.46 (2.45) & 31.33 (15.58) & 4.54 (1.11) & 2.24 (0.57) & 270.44 (78.31) & [-0.18, 1.01] \\
    4 & 79.98 (34.09) & 0.96 (3.43) & 27.92 (15.82) & 3.82 (1.33) & 1.79 (0.76) & 218.76 (95.23) & [1.01, 2.20] \\ 
    \hline
    \end{tabular}}%
    \end{center}
    \label{RulesGENNTSKwRLS}%
\end{table}%

\subsection{Discussion}

Regarding the proposed models, NTSK with feature selection approach achieved superior performance compared to the other proposed models. This suggests that feature selection is effective in addressing the challenges posed by noisy datasets. For eFSs, the eTS and exTS models outperformed the rest of eFSs. Notably, Simpl\_eTS generated the highest number of rules in many simulations. Regarding DL models, CNN exhibited superior performance for the renewable energy datasets, achieving the lowest errors. Finally, among the classical models, LS-SVM achieved the lowest errors for the renewable energy datasets. 

The proposed approach for designing fuzzy rules offers two key advantages: simplicity and fewer hyperparameters. This simplicity translates to reduced training and testing times, making the proposed models well-suited for real-world applications requiring rapid implementation. Despite their simplicity and non-evolving structure, the proposed models achieved comparable or even superior error results compared to eFSs. Another notable advantage of the proposed models is that users can directly specify the desired number of rules, unlike most rule-based models in the literature. This feature allows users to customize the number of rules to balance accuracy and interpretability. Furthermore, the implementation of a genetic algorithm for attribute selection enhances the models’ robustness, enabling them to handle diverse datasets more effectively by reducing errors and improving interpretability. Lastly, the ensemble approach contributes to reducing uncertainty in the results.

\section{Conclusion}\label{conclusions}

This work introduces new data-driven fuzzy inference systems (NFISiS) for time series forecasting. NFISiS includes a series of fuzzy models, including: the new Mamdani classifier (NMC), new Mamdani regressor (NMR) and the new Takagi-Sugeno-Kang (NTSK), GEN-NMR, GEN-NTSK, R-NMR, R-NTSK. For the NTSK, two adaptive filtering are implemented, the Recursive Least Squares (RLS) and the weighted Recursive Least Squares (wRLS), producing the NTSK (RLS) and NTSK (wRLS). The following advantages of the proposed model can be highlighted: (i) a novel data-driven mechanism for designing Mamdani and TSK rules that provides reduced complexity, higher autonomy and accuracy, and less hyperparameters; (ii) the implementation of feature selection approaches that enhances the models’ ability to
handle large datasets, optimize their performance, increase interpretability, and avoid
overfitting. The models are applied solar energy datasets. Their performance are evaluated in terms of errors and the number of rules. GEN-NTSK and RF-NTSK obtained the lowest error among the proposed models.

One of the main advantages of NTSK concerning the conventional TSK is their increased interpretability, as the polynomial functions of the consequent part can be replaced by the expected variation of the target value when presenting the rule-based structure. As the demand for mechanisms ensuring the safe operation of critical systems continues to rise, the models proposed in this work are well-suited for implementation as control tools due to their reliability and simplicity. Their application can support decision-making across various domains, such as finance, by improving system performance and optimizing monetary outcomes. For future research, we recommend exploring alternative fuzzy set types and developing a method for automatically selecting the optimal fuzzy set shape. Additionally, a post-processing stage for fine-tuning fuzzy set parameters could be investigated. Other algorithms, such as kernel recursive least squares (KRLS), may also be explored for estimating the parameters of the consequent part. Finally, we suggest further evaluating the proposed models on diverse datasets, including those with higher dimensionality and increased uncertainty, to compare their performance with type-2 fuzzy sets.

\bibliographystyle{unsrtnat}
%\bibliography{paper}  %%% Uncomment this line and comment out the ``thebibliography'' section below to use the external .bib file (using bibtex) .


\end{document}